\documentclass[runningheads]{llncs}
\usepackage[T1]{fontenc}
\usepackage{graphicx}
\usepackage{graphicx}
\usepackage{amsmath}
\usepackage{booktabs}
\usepackage[subtle]{savetrees}
\usepackage{comment}

\usepackage{subcaption}
\usepackage{float}
\usepackage{url}
\usepackage{xurl}
\usepackage{hyperref}
\begin{document}
\title{Bridging Forecast Accuracy and Inventory KPIs: A Simulation-Based Software Framework}
\titlerunning{Bridging Forecast Accuracy and Inventory KPIs}
%
%
\author{So Fukuhara\inst{1,2}\orcidID{0009-0001-2307-4657} \and\\
Abdallah Alabdallah\inst{1}\orcidID{0000-0001-9416-5647} \and
Nuwan Gunasekara\inst{1}\orcidID{0000-0002-7964-6036} \and
S\l{}awomir Nowaczyk\inst{1}\orcidID{0000-0002-7796-5201}}
\authorrunning{S. Fukuhara et al.}
%
\institute{Center for Applied Intelligent Systems Research, Halmstad University, Sweden. \\
\email{\{sofuk25,abdallah.alabdallah,nuwan.gunasekara,slawomir.nowaczyk\}@hh.se}\\ \and
Kagawa University, Kagawa, Japan\\
\email{s24d164@kagawa-u.ac.jp}}
\maketitle              
\begin{abstract}
Efficient management of spare parts inventory is crucial in the automotive aftermarket, where demand is highly intermittent and uncertainty drives substantial cost and service risks. Forecasting is therefore central, but the quality of a forecasting model should be judged not by statistical accuracy (e.g., MAE, RMSE, IAE) but rather by its impact on key operational performance indicators (KPIs), such as total cost and service level. Yet most existing work evaluates models exclusively using accuracy metrics, and the relationship between these metrics and operational KPIs remains poorly understood.

To address this gap, we propose a decision-centric simulation software framework that enables the systematic evaluation of forecasting models in realistic inventory management settings. The framework comprises: (i) a synthetic demand generator tailored to spare-parts demand characteristics, (ii) a flexible forecasting module that can host arbitrary predictive models, and (iii) an inventory control simulator that consumes the forecasts and computes, based on selected inventory control policy, operational KPIs. This closed-loop setup enables practitioners and researchers to evaluate models not only in terms of statistical error but also in terms of their downstream implications for inventory decisions.

Using a wide range of simulation scenarios, we show that improvements in conventional accuracy metrics do not necessarily translate into better operational performance, and that models with similar statistical error profiles can induce markedly different cost–service trade-offs. We analyze these discrepancies to characterize how specific aspects of forecast performance affect inventory outcomes and to derive actionable guidance for model selection. Overall, the framework operationalizes the link between demand forecasting and inventory management, shifting evaluation from purely predictive accuracy towards operational relevance in the automotive aftermarket and related domains.

An open-source implementation of the software, including all experimental results, is available at \url{https://github.com/caisr-hh/TruckParts-Demand-Inventory-Simulator/releases/tag/IDA_2026}.
\end{abstract}
\section{Introduction}
The automotive aftermarket plays a central role in the vehicle life cycle, ensuring the availability of spare parts for maintenance and repair over extended periods. Efficient management of inventory is therefore crucial, both economically and environmentally. Overstocking ties up capital and leads to scrapping due to obsolescence, while understocking increases downtime, emergency shipments, and customer dissatisfaction. These trade-offs are particularly acute for rare and expensive components, where demand is sparse and highly uncertain, and for parts associated with emerging technologies such as electric-vehicle traction batteries. Limited supply, short shelf lives, rapid product evolution, and limited operational experience all contribute to the complexity of related logistics processes.

From a sustainability perspective, improving spare parts logistics offers substantial leverage. Better forecasts and more informed inventory decisions can reduce the number of components produced but never used, lower the frequency of rush orders, and shift transport away from fast but environmentally costly modes (e.g., air) towards slower but greener alternatives (e.g., sea or consolidated road shipments). At the same time, avoiding unnecessary back-and-forth movements of parts between warehouses, regional hubs, and dealers can substantially decrease the overall environmental footprint of vehicle operation. Achieving ``the right part in the right place at the right time'' is thus not only an economic objective, but also a key enabler for more resource-efficient and sustainable transport.

However, accurately predicting spare parts demand remains a major challenge. Demand is often intermittent or sporadic, with signals about upcoming needs being delayed or incomplete. Parts may be phased out or replaced with little notice, and supply chain capacity is often constrained. On a large scale, even determining the current life cycle stage of parts in the field is challenging. In practice, inventory policies at warehouses and dealers are still frequently based on coarse-grained rules, with manual and ad hoc adjustments made to account for local factors. This leads to suboptimal resource utilization, making it difficult to systematically assess the impact of new forecasting methods or data sources.

Recent advances in AI and ML offer new opportunities to address these issues. Richer data, including predictive maintenance signals, telematics, and warranty information, can provide earlier and more granular indications of future demand. At the same time, modern forecasting approaches, such as meta-learning and other data-driven methods that leverage information across large families of time series while adapting to local idiosyncrasies, are particularly well-suited to the heterogeneous and sparse nature of spare parts demand. Yet, despite substantial progress in predictive modeling, the evaluation of such models is still dominated by statistical accuracy metrics, with a limited understanding of how changes in these metrics translate into operational performance in the real world.

This paper addresses the interface between demand forecasting and inventory management. We develop and study a decision-centric simulation framework that links forecasting models to downstream inventory decisions under realistic constraints, clarifying how different aspects of forecast performance shape cost–service trade-offs and highlighting how common metrics can be misleading. 
%
%
\section{Spare Parts Forecasting Methods}
Spare parts forecasting methods are broadly classified into two categories: time-series methods, which rely on historical demand data, and contextual methods, which incorporate additional information, such as expert judgment or equipment data \cite{pincce2021intermittent}. Classical methods, such as Simple Exponential Smoothing (SES) and ARIMA, are often unsuitable for spare parts due to their erratic, lumpy, or intermittent demand patterns \cite{pincce2021intermittent}, making it common to use ML methods such as Random Forests, SVMs, or ANNs.

Time-series methods use historical demand data and can be parametric, nonparametric, or employ improvement strategies. Parametric approaches assume lead-time demand follows a known probability distribution. Croston's method separately forecasts demand intervals and sizes \cite{croston1972forecasting}, later improved by the Syntetos-Boylan Approximation (SBA) with a bias correction term \cite{syntetos2005accuracy}. To address demand obsolescence, the TSB method incorporates demand probability forecasts, which are updated every period \cite{teunter2011}. Parametric bootstrapping methods simulate lead-time demand using estimated parameters but can underestimate variability, potentially leading to insufficient safety stock \cite{snyder2002forecasting,hasni2019performance}. In contrast, nonparametric approaches reconstruct the empirical demand distribution without assuming a specific form, typically via bootstrapping. The method by \cite{willemain2004new} utilizes a Markov process for demand occurrence and performs well for highly lumpy demand, whereas neural networks excel at capturing nonlinear patterns, such as intermittence \cite{gutierrez2008,kourentzes2013intermittent}. Forecasts can be further improved through strategies like demand classification, which matches patterns with optimal methods \cite{syntetos2005categorization}, and data aggregation, which reduces variability from zero-demand periods to create more stable forecasts \cite{nikolopoulos2011aggregate}.

Contextual forecasting integrates historical data with external information. Judgmental forecasting adjusts statistical forecasts with expert knowledge, where systematic, particularly negative, adjustments can improve intermittent demand forecasts; however, the effectiveness of this approach varies \cite{syntetos2009effects}. Installed base forecasting uses information about the operational equipment portfolio, incorporating data on quantity, age, and maintenance schedules, proving particularly effective for highly intermittent demand in B2B settings \cite{dekker2013use,hua2007new,zhu2020spare}, despite challenges with data collection costs \cite{andersson2018big}.

This paper is motivated and focuses on heavy-duty vehicle aftermarket logistics; however, our overall software framework is flexible and can also be applied to other domains by adapting relevant parameters.
\section{Spare Parts Demand Data Generator}
Synthetic data provides a practical alternative to real operational records, 
avoiding confidentiality constraints and offering a common evaluation environment. 
It can be generated at any scale or structure, enabling the testing of diverse scenarios. 
This study develops a synthetic demand generator for aftermarket truck parts.
The simulator replicates truck operating structure, usage conditions, and seasonal factors, and generates demand time series by simulating part failures using a survival-model.

We are not permitted to share real-world demand data from our industrial partner, but generating synthetic data with similar characteristics allows us to share our results with the scientific community and promotes reproducible research.
\subsection{Concept and Structure}
The simulator uses a Discrete Time Simulation (DTS) approach in which time advances in fixed increments (in this study, one day) and the system state is updated at each increment. 
\begin{figure}[tb]
    \centering
    \includegraphics[width=0.8\linewidth]{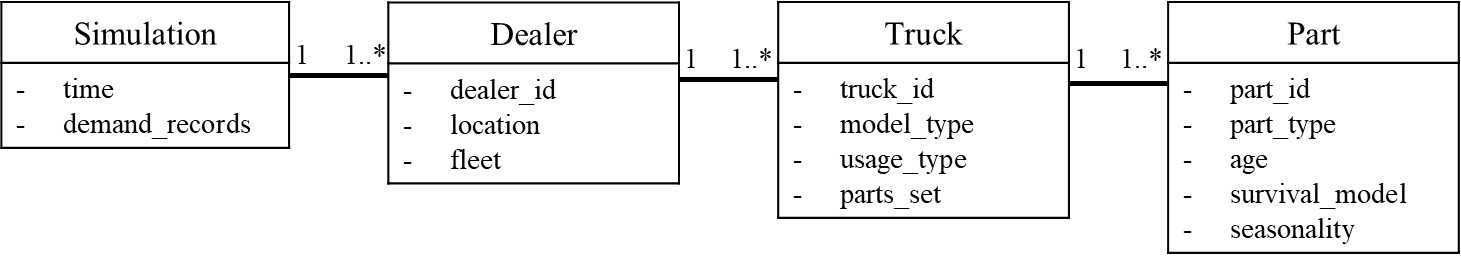}
    \caption{Class Diagram.}
    \label{fig:class_diagram}
\end{figure}
The class diagram of the simulator is shown in Figure \ref{fig:class_diagram}, employing a hierarchical structure 
\texttt{Dealer} $\to$ \texttt{Truck} $\to$ \texttt{Part}, as described below:\\
• The \texttt{Simulation} class manages the overall system, maintaining the current system time (simulation date) and creating a log of demand occurrences;\\
• Multiple \texttt{Dealer} objects are instantiated, with unique identifiers and location information. Each dealer manages a collection of trucks;\\
• Each \texttt{Truck} is identified by a model type and usage type;\\
• Finally, every truck is composed of a set of \texttt{Part} objects. Parts belong to a specific type (an abstract category, e.g., engine), and trucks of the same vehicle model contain an identical set of part types.
Part types have a specific ``baseline'' survival model and are affected by seasonality and usage in predetermined ways.
A part instance, on the other hand, keeps track of its own age (a broken part gets replaced with a new one).
\subsection{Survival Model}
The simulator uses survival models to probabilistically simulate spare-parts demand. 
We have implemented survival models widely used in reliability engineering, treating time-to-event as a random variable and describing its distribution together with the effects of covariates \cite{Karim2019} \cite{Gorjian2010}.

In the simulator, part failure is the event of interest. 
For any part $p$, the failure probability is evaluated conditionally on the time elapsed since its last failure, denoted as $t_p$.
Specifically, for a part that has survived until time $t_p$, 
the probability of failure within the time interval $[t_p,\, t_p+\Delta t]$ is given by
\begin{equation}
    \Pr\left(t_p < T_p \le t_p + \Delta t \mid T_p > t_p \right)
    = 1 - \exp\!\left( - \int_{t_p}^{t_p+\Delta t} h(u)\,du \right),
    \label{eq:failure_prob}
\end{equation}
where $\Delta t$ is the simulation time-step and $T_p$ is the random variable for time to failure. 
$h(t)$ denotes the hazard function, defined as the instantaneous failure rate at $t$.
Hazard functions have been extensively studied to capture a wide range of failure behaviors. Four parametric models, $\{\texttt{Exponential}, \texttt{Weibull}, \texttt{Log-logistic}, \texttt{Gompertz}\}$, are employed.
These models exhibit distinct temporal hazard behaviors, enabling heterogeneous failure dynamics across parts. 
Their definitions and parameter settings, conditioned on the truck's usage style, are presented below. The \texttt{Hard} setting indicates severe operating conditions with higher and earlier failure risk, whereas \texttt{Normal} represents standard usage conditions.

\newcommand{\sss}[1]{\vspace{2mm}\noindent\textbf{#1}. }

\sss{Exponential Model}
The hazard function is defined as
\begin{equation}
    h(t) = \lambda,\qquad t \ge 0,
\end{equation}
where $\lambda$ is the model parameter. This constant rate signifies that the probability of failure remains the same at any point in time, reflecting the ``memoryless'' property, which means past events do not influence the likelihood of future events.
It determines the median failure time $t_m$, at which the probability of failure reaches 50\%, as follows:
\begin{equation}
    t_m = \frac{\ln(2)}{\lambda}.
    \label{eq:tm1}
\end{equation}
From a user perspective, $t_m$ is a more interpretable parameter than $\lambda$.
Accordingly, the simulator accepts a range of values for $t_m$ as a configuration parameter, selects one uniformly at random for each part type, and computes $\lambda$ accordingly.


\sss{Weibull Model}
The hazard function is defined as
\begin{equation}
    h(t) = \frac{k}{\lambda} \left(\frac{t}{\lambda}\right)^{k-1},\qquad t \ge 0,
\end{equation}
where $\lambda$ and $k$ are the scale and shape parameters, respectively, and $t_m$ is:
\begin{equation}
t_m = \lambda \left(\ln 2\right)^{1/k}.
\label{eq:tm2}
\end{equation}
The simulator first assigns the value to $k$ randomly (within a given range), and then (as above), calculates $\lambda$ according to the user-specified desired median failure time $t_m$.

\sss{Log-logistic Model}
The hazard function and median failure time $t_m$ are defined as
\begin{equation}
    h(t) = \frac{k \lambda \, t^{k-1}}{1 + (\lambda t)^k}, 
    \qquad t > 0,
    \qquad t_m = \lambda^{-1} (\ln 2)^{1/k}.
    \label{eq:tm3}
\end{equation}

\sss{Gompertz Model}
The hazard function and median failure time $t_m$ are defined as
\begin{equation}
    h(t) = k \lambda e^{\lambda t},
    \qquad t > 0,
    \qquad t_m = \frac{1}{\lambda} 
      \ln\!\left( 1 + \frac{\ln 2}{k} \right).
      \label{eq:tm4}
\end{equation}

\subsection{Seasonality}
Seasonal conditions significantly impact failure probabilities, and the nature of this influence varies by component type and operating environment. For example, engine-related parts tend to exhibit higher failure rates during cold periods, driven by increased cold starts and additional load on auxiliary heating systems. In contrast, tire components are sensitive to road-surface conditions associated with freeze--thaw cycles in winter and to temperature-induced rubber deterioration during summer. 
\begin{figure}[h]
    \centering
        \includegraphics[scale=0.3]{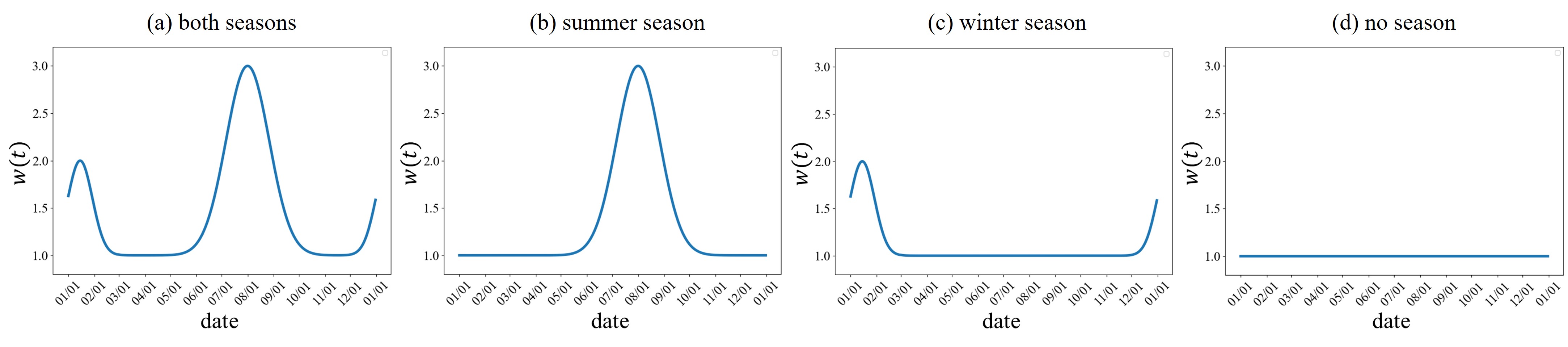}
    \caption{Curves of seasonality coefficients $w(t)$ for the period of 12 months (Jan-Dec): a) both seasons, b) summer, c) winter, d) no seasonality.
    }
    \label{fig:season_coef}
\end{figure}
Moreover, regional climate further reinforces these effects: prolonged and harsher winters in cold regions amplify seasonal stresses on components, whereas high heat and humidity in hot regions accelerate mechanical degradation. We incorporate these seasonal effects into the hazard function, focusing on two seasons, summer and winter, and two regional types, southern and northern. Simulator introduces a seasonal coefficient $w(t)$ for each part, multiplying the baseline hazard function $h(t)$ as
\begin{equation}
    h_{\text{seasonal}}(t) = w(t) \cdot h(t),
\end{equation}
and $w(t)$, capturing seasonal fluctuations, is modelled using a Radial Basis Function:
\begin{equation}
    w(t) = \sum_{m=1}^{M} 
    A_m \exp\!\left(
        -\frac{
            \left\|
                \max\!\left(
                    |\mu_m - d(t)|,\,
                    D - |\mu_m - d(t)|
                \right)
            \right\|^2
        }{\sigma_m^2}
    \right),
    \label{eq:rbf}
\end{equation}
where $M$ denotes the number of basis functions, and $m=1,\dots,M$ indexes  each basis component; $t$ denotes the cumulative simulation time 
and is mapped to the day-of-year via $d(t)$, 
with the length of one year $D$ (typically $D=365$).
The parameters $\mu_m$, $\sigma_m$, and $A_m$ denote the center, width, and amplitude of the $m$-th basis function, controlling the timing, duration, and strength of seasonal effects.
To ensure annual periodicity, the model employs the modular distance
$\max\!\left(|\mu_m - d(t)|,\; D - |\mu_m - d(t)|\right)$~\cite{duvenaud2014kernel}.


This formulation allows the seasonal coefficient $w(t)$ to flexibly represent region- and part-specific patterns. Examples of the patterns used in our experiments can be seen in Figure~\ref{fig:season_coef}.
\subsection{Concept Drift}
In practice, the number of trucks managed by a dealer often changes over time due to operational growth, demand fluctuations, or business decisions. The simulator models two types of concept drifts: sudden (the number of trucks for a dealer increases abruptly, for example, from 50 to 75 trucks, a dealer-specific random time point) and slow (the number of trucks increases gradually, for example, by one truck per week).
\subsection{Simulation}
\subsubsection{Input Data}
The main input parameters of the simulation are as follows:
\renewcommand{\labelitemi}{\textbullet}
\begin{itemize}
    \item \textbf{Simulation start date}: denoted by $T_0$
    \item \textbf{Simulation horizon} (days): denoted by $T$.
    \item \textbf{Time-step size}: denoted by $\Delta t$.
    \item \textbf{Number of dealers}: denoted by $N_d$.
    \item \textbf{Number of trucks per dealer (range)}: 
    $N_t \in [N_t^{(l)},\, N_t^{(u)}]$.
    \item \textbf{Number of part types constituting a truck (range)}:  
    $N_p \in [N_p^{(l)},\, N_p^{(u)}]$.
    \item \textbf{Seasonality parameters}:
    the RBF parameters
    $ ( \boldsymbol{\mu}, \boldsymbol{\sigma}, \boldsymbol{A} )$
    in equation~(\ref{eq:rbf}).
    \item \textbf{Parameters of survival models}:  
    the range of the median time $t_m \in [t_m^{(l)},\, t_m^{(u)}] $ 
    in equations~\ref{eq:tm1}, \ref{eq:tm2}, \ref{eq:tm3} and \ref{eq:tm4}.
\end{itemize}

\sss{Process Flow}
The simulation proceeds according to the following steps:
\begin{enumerate}
    \item \textbf{Initialization}:
    The simulation is initialized as follows:   
    \begin{enumerate}
        \item The simulation time is initialized as 
        \( \texttt{time} \leftarrow 0\).
        
        \item A total of \(N_d\) \texttt{Dealer}\ instances are created. For each dealer:
        \begin{itemize}
            \item Assign a $\texttt{location}$ from 
            $\{\texttt{southern},\, \texttt{northern}\}$.
            \item Sample the beginning time of sudden drift, 
            $
                 \texttt{sudden\_time} \in  [T_0,\, T_0 + T].
            $

            \item Sample the beginning time of slow drift,
            $
                 \texttt{slow\_time} \in  [T_0,\, T_0 + T].
            $
                        
            \item Sample the number of trucks managed by the dealer from 
            $N_t \in [N_t^{(l)},\, N_t^{(u)}]$.

            \item Assign a \texttt{survival\_model} and \texttt{seasonality} category to each part type.
        \end{itemize}
        
        \item A total of \(N_t\) \texttt{Truck} instances are generated. 
        \begin{itemize}
            \item Each truck is assigned a \texttt{usage} from $\{\texttt{Normal},\, \texttt{Hard}\}$.
            \item Each part in a truck is assigned $\texttt{age} \leftarrow 0$.
        \end{itemize}



        
    \end{enumerate}

    \item \textbf{Demand Evaluation}:     
    At any given time step, the simulator first evaluates which parts of the trucks failed, as follows:

    \begin{enumerate}
        \item \textit{Failure probability: }
        The probability that part \(p\) fails within the interval 
        \([t_p,\, t_p + \Delta t]\), given that it has survived up to time \(t_p\) (\texttt{age}), is computed as  
        \[
            \Pr(t_p < T \le t_p + \Delta t \mid T > t_p).
        \]
    
        \item \textit{Failure occurrence decision:}  
        A failure is triggered when the probability exceeds a uniform random draw:
        \begin{equation}
            \text{Demand} =
            \begin{cases}
            \text{True}, &
                 \text{if }
                 \Pr(t_p < T \le t_p+\Delta t_p \mid T > t_p) \ge U(0,1), \\[6pt]
            \text{False}, & \text{otherwise}.
            \end{cases}
        \end{equation}
    
        \item \textit{Post-failure processing:}  
        When a failure occurs, the simulator records the event  
        (\texttt{time}, \texttt{dealer\_id}, \texttt{truck\_id}, 
        \texttt{part\_id}, \texttt{part\_type})  
        and resets \(\texttt{age} \leftarrow 0\).
    \end{enumerate}

    \item \textbf{Concept Drift}:     
    The simulator, for each dealer, evaluates:
    \begin{enumerate}
        \item \textit{sudden drift}: When \texttt{time} reaches \texttt{sudden\_time}, the dealer generates additional \texttt{Truck} instances: \( N_t \leftarrow \alpha \cdot N_t ,\qquad \alpha \in \mathrm{rand}(1.25,\, 1.5). \)
        \item \textit{slow drift}: If $\texttt{time} \geq \texttt{slow\_time}$, the dealer experiences slow drift and generates new \texttt{Truck} instance \( N_t \leftarrow N_t + 1 \).
    \end{enumerate}

    \item \textbf{Time Update and Termination}:
    The simulation time is incremented as \(\texttt{time} \leftarrow \texttt{time} + \Delta t\), and the age of each part is updated accordingly, \(\texttt{age} \leftarrow \texttt{age} + \Delta t\). If $\texttt{time} == T$, the simulation terminates; otherwise, it returns to Step~2.
    
    \item \textbf{Output}:    
    Upon termination, daily demand counts are aggregated for each part type within each dealer.  
    The simulator outputs a set of daily demand time series data for all  
    \((\texttt{dealer\_id},\ \texttt{part\_type})\) combinations.
    \end{enumerate}
\section{Cost Simulator}
To simulate the cost, we utilize the cost simulator \cite{kamil2025prediction}, which employs Discrete-Event Simulation (DES) \cite{banks2005discrete}, a technique previously applied to supply chain planning \cite{hellstrom2020simulation}. The simulator models the inventory management process for individual parts at dealers, incorporating stochastic elements such as demand variability and fixed lead times. It operates through three primary event types: demand (customer orders), delivery (replenishment orders arriving), and inventory check (periodic reviews triggering reorders). The simulator \cite{kamil2025prediction} uses a ``standard inventory policy'' based on normally distributed demand assumptions, calculating safety stock, reorder points, and order quantities using established formulas \cite{banks2005discrete,hellstrom2020simulation}.

\begin{figure}[tb]
    \centering
    \includegraphics[width=0.98\linewidth]{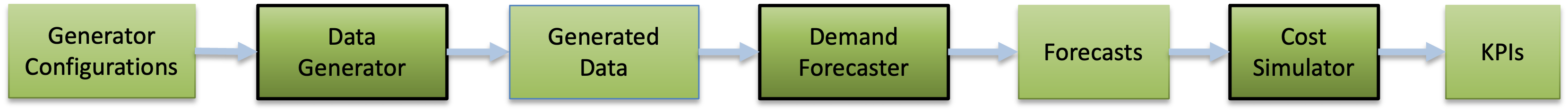}
    \caption{Overall flow of the software framework}
    \label{fig:SimulationFlow}
\end{figure}

The cost simulator incorporates multiple cost components to provide a comprehensive evaluation of inventory strategies, including holding costs, order costs, rush order costs, transportation costs, and badwill costs (penalties for service level shortfalls). Inputs to the model include historical and forecasted demand data, lead times, and inventory policy parameters. In this work, we maintain constant inventory policy parameters for all parts across all dealers. Output metrics focus on KPIs such as \textit{total cost} and \textit{service level} (fill rate), allowing for a direct comparison of different forecasting approaches under controlled conditions. Figure \ref{fig:SimulationFlow} shows this setup as a flow diagram. 

\section{Experiments}
We created 48 synthetic demand scenarios of three years of synthetic data, forecasted demand using several models, and evaluated corresponding business KPIs.
Four concept-drift conditions were considered: without drift, sudden drift only, slow drift only, and combined drift.
For each drift condition, we tested twelve parameter configurations constructed by combining four ranges of the number of trucks per dealer, 
$N_t \in {[5,\,10], [10,\,30], [30,\,50], [50,\,100]} $, 
with three ranges of median lifetime,
$t_m \in {[100,\, 150], [150,\, 365], [365,\, 730]} $ days. 
The remaining settings were fixed, with the simulation horizon $T = 1095$ days, the simulation step $\Delta t = 1$ day, and four dealers ($N_d = 4$).

For forecasting, we trained models on the first 2 years of demand data and forecasted demand for the following year. 
We employed three machine-learning models: XGBoost, Random Forest, and SVR, and four statistical time-series forecasting methods: ARIMA, Croston's method, TSB, and SBA. Hyperparameters for each model were tuned to minimize MAE. 

The Cost Simulator evaluates total inventory-management costs for each part type at each dealer over a one-year period.
The following simplifying assumptions were made. All parts share the same price, unit cost, weight, initial inventory, and lead time. A single supplier with unlimited capacity. No partial shipments, cancellations, or modifications to orders. Orders are processed independently, and batching effects are ignored.

\begin{figure}[t]
    \centering
    \includegraphics[width=1.0\linewidth]{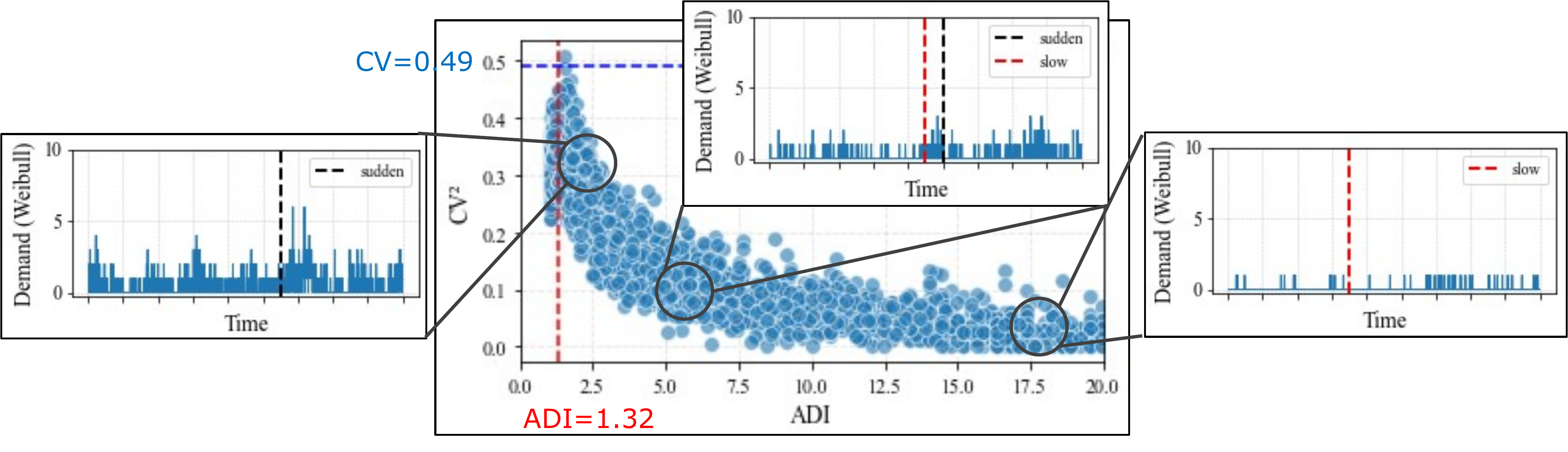}
    \caption{ADI--CV$^{2}$ distribution for all demand series, with examples.} 
    \label{fig:adi_cv2}
\end{figure}

\sss{Demand Generation Results}
We present two metrics for the generated demand: 
the Average Demand Interval (ADI) and the squared Coefficient of Variation (\(\text{CV}^2\)).
ADI quantifies the demand intermittency by measuring the average interval between non-zero demands, 
whereas \(\text{CV}^2\) captures demand variability.
%
Figure~\ref{fig:adi_cv2} presents the ADI--CV$^{2}$ distribution of all generated demand series across all scenarios.
The insets display representative demand series of selected parts: 
the first one has summer and winter seasonality with a sudden drift, 
while the second and third include only combined and slow drift, respectively.
The majority of generated series fall within the intermittent region, confirming that the generator successfully produces real-world-like intermittent demand, with substantial variety in behavior.

\sss{Forecasting Results}
We evaluated the forecasting performance using four metrics:  
the Mean Absolute Error (MAE), the Root Mean Squared Error (RMSE), the coefficient of determination ($R^{2}$),
and the Intermittent Accuracy Error (IAE)~\cite{Flyckt1991114}, which is designed to assess accuracy for highly intermittent demand (originally created for make-to-order manufacturing, which shares a lot of similarity with aftermarket logistics).  
Table~\ref{tab:metric_cost} presents the mean and standard deviation of each metric across all scenarios, with the best value in each metric in bold. XGBoost and Random Forest 
achieved the best MAE, $R^{2}$, and IAE; Random Forest obtained the lowest MAE, RMSE, and IAE. 
In terms of \(R^2\), all models show negative values, even when accounting for standard deviation, implying the demand series are highly irregular and challenging to predict.

\begin{table}[tb]
    \centering
    \caption{Prediction accuracy metrics and total inventory cost for each forecasting model (bold indicates the best performing model, per metric).}
    \label{tab:metric_cost}
    \begin{tabular}{lccccc}
    \toprule
    Model & MAE & RMSE & $R^2$ & IAE & Total Cost \\
    \midrule
    XGBoost      & \textbf{0.43} $\pm$ 0.36 & 0.60 $\pm$ 0.43 &  \textbf{0.02} $\pm$ 0.11 & \textbf{0.80} $\pm$ 0.19 & $8.2\times10^5$ \\
    RandomForest & \textbf{0.43} $\pm$ 0.35 & \textbf{0.59} $\pm$ 0.42 & 0.01 $\pm$ 0.10 & \textbf{0.80} $\pm$ 0.19 & $7.5\times10^5$ \\
    SVR          & 0.50 $\pm$ 0.35 & 0.64 $\pm$ 0.46 & -0.34 $\pm$ 1.05 & 0.84 $\pm$ 0.20 & $8.4\times10^5$ \\
    ARIMA        & 0.53 $\pm$ 0.57 & 0.68 $\pm$ 0.62 & -0.12 $\pm$ 0.28 & 0.81 $\pm$ 0.16 & $7.5\times10^5$ \\
    CROSTON      & 0.52 $\pm$ 0.56 & 0.67 $\pm$ 0.62 & -0.10 $\pm$ 0.25 & \textbf{0.80} $\pm$ 0.16 & $\mathbf{7.4\times10^5}$ \\
    SBA          & 0.51 $\pm$ 0.53 & 0.66 $\pm$ 0.59 & -0.08 $\pm$ 0.20 & \textbf{0.80} $\pm$ 0.16 & $7.8\times10^5$ \\
    TSB          & 0.57 $\pm$ 0.63 & 0.73 $\pm$ 0.68 & -0.34 $\pm$ 0.71 & 0.83 $\pm$ 0.16 & $7.8\times10^5$ \\
    \bottomrule
\end{tabular}
\end{table}

\begin{figure}[t]
    \centering
    \includegraphics[width=0.8\linewidth]{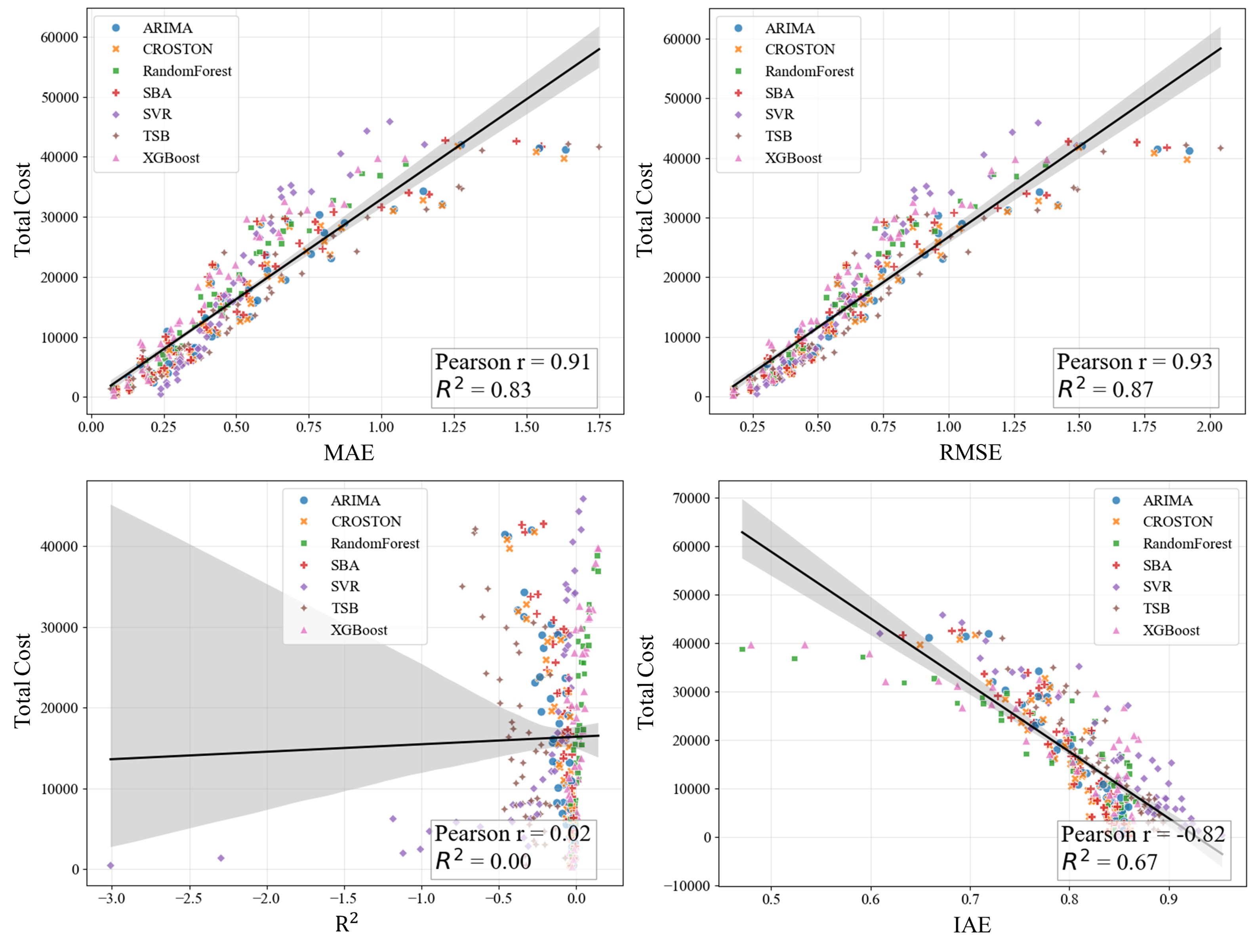}
    \caption{Comparison between total cost and each evaluation metric.}
    \label{fig:cost_metric_1}
\end{figure}

\sss{Cost Simulation Results}
Table~\ref{tab:metric_cost} also presents the total inventory-management cost for each forecasting model across all scenarios. The lowest total cost is achieved by Croston’s method, followed by Random Forest, ARIMA, SBA, and TSB. In contrast, XGBoost and SVR incur substantially higher costs than the other models. This pattern stands in contrast to the forecasting accuracy in the same table, where Croston, SBA, TSB, and ARIMA perform markedly worse than XGBoost. This demonstrates that higher forecasting accuracy does not necessarily result in lower operational costs, as seen in Figure~\ref{fig:cost_metric_1}.

\begin{figure}[t]
\centering
\begin{subfigure}{0.50\columnwidth}
  \centering
  \includegraphics[width=1\linewidth]{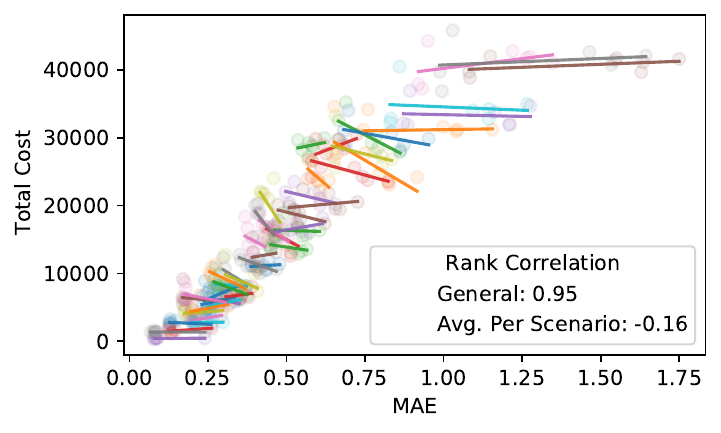}
  \caption{MAE}
  \label{fig:cost_vs_MAE}
\end{subfigure}%
%
~
\begin{subfigure}{0.50\columnwidth}
  \centering
  \includegraphics[width=1\linewidth]{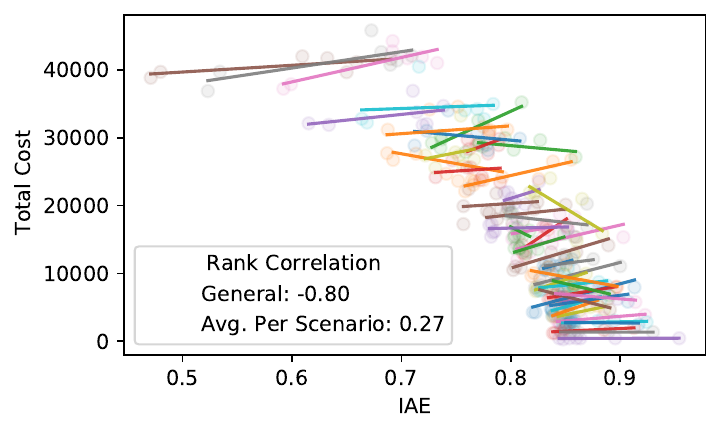}
  \caption{IAE}
  \label{fig:cost_vs_IAE}
\end{subfigure}
\caption{Scenario comparison between total cost and each evaluation metric.}
\label{fig:cost_metric_agg}
\end{figure}

Figure~\ref{fig:cost_metric_agg} shows the relationship between forecasting metrics and the total inventory cost across multiple simulated scenarios. Each point represents a model’s mean metric and cost under a given scenario, with colors distinguishing scenarios, and the lines show regression fits for each scenario. 
One can clearly see a counterintuitive pattern known as Simpson's paradox~\cite{Neufeld1995}. The per-scenario correlations, reflected in the slopes of the regression lines, vary widely and are predominantly negative, averaging -0.16 and -0.06 for MAE and RMSE, respectively, indicating that lower MAE or RMSE does not imply lower cost.

\section{Conclusions}

The goal of this work was to bridge the persistent gap between forecast evaluation based on statistical accuracy and evaluation based on operational performance in spare parts inventory management. To this end, we developed a decision-centric simulation framework that couples a synthetic demand generator, a flexible forecasting module, and an inventory cost simulator under realistic operating assumptions. The framework is implemented as open-source software, together with all experimental configurations, thereby providing a reusable testbed for transparent and reproducible research on demand forecasting and inventory control in the automotive aftermarket and related domains. 

Our empirical study across a wide range of scenarios demonstrates that conventional accuracy metrics (MAE, RMSE, R2, IAE) are only weakly aligned with business-level outcomes. Models that achieve the best statistical performance (XGBoost and Random Forest) do not necessarily minimize total inventory cost. In contrast, classical intermittent-demand methods, including Croston’s, SBA, and TSB, can yield substantially lower costs despite lower accuracy scores. By jointly analyzing accuracy metrics and cost-based KPIs, we can quantify this discrepancy and show that model rankings can change dramatically once service levels, rush orders, and other cost components are taken into account. These findings highlight the risk of relying solely on statistical error measures when selecting forecasting models for real inventory systems.

Taken together, the two main contributions of this paper are: (i) a modular, open-source software framework that lowers the barrier to conducting rigorous, reproducible experiments at the interface of forecasting and inventory management; and (ii) an empirical quantification of how and when improvements in accuracy metrics fail to translate into better cost–service trade-offs. This opens several avenues for future work, including extending the framework to multi-echelon networks and richer cost structures, integrating more advanced learning-based forecasters, and calibrating the simulator to multiple real-world case studies. Ultimately, we hope that the proposed framework will encourage the community to adopt decision-centric evaluation protocols that prioritize operational relevance over purely predictive accuracy, making it the central focus of methodological development.

\bibliographystyle{splncs04}
\bibliography{mybibliography}
%




\end{document}